\documentclass[]{article}
\usepackage{preprint}

\usepackage[utf8]{inputenc}	% allow utf-8 input
\usepackage{doi}

\usepackage[numbers]{natbib}
\usepackage{multicol}
\usepackage{graphicx}

\usepackage{physics}
\usepackage{amsmath}
\usepackage{tikz}
\usepackage{mathdots}
\usepackage{yhmath}
\usepackage{cancel}
\usepackage{color}
\usepackage{siunitx}
\usepackage{array}
\usepackage{multirow}
\usepackage{amssymb}
\usepackage{gensymb}
\usepackage{tabularx}
\usepackage{extarrows}
\usepackage{booktabs}
\usepackage{float}
\usetikzlibrary{fadings}
\usetikzlibrary{patterns}
\usetikzlibrary{shadows.blur}
\usetikzlibrary{shapes}

\usepackage{cleveref}
\usepackage[skip=3pt]{caption}

\graphicspath{{pictures/}}

\usepackage{newfloat}
\DeclareFloatingEnvironment[name={Supplementary Figure}]{suppfigure}
\usepackage{sidecap}
\sidecaptionvpos{figure}{c}

\usepackage{titlesec}
\titlespacing\section{0pt}{12pt plus 3pt minus 3pt}{1pt plus 1pt minus 1pt}
\titlespacing\subsection{0pt}{10pt plus 3pt minus 3pt}{1pt plus 1pt minus 1pt}
\titlespacing\subsubsection{0pt}{8pt plus 3pt minus 3pt}{1pt plus 1pt minus 1pt}

\title{S-Nav: Semantic-Geometric Planning for Mobile Robots}

\usepackage{tikz,xcolor,hyperref}

\definecolor{lime}{HTML}{A6CE39}
\DeclareRobustCommand{\orcidicon}{
	\begin{tikzpicture}
	\draw[lime, fill=lime] (0,0) 
	circle [radius=0.16] 
	node[white] {{\fontfamily{qag}\selectfont \tiny ID}};
	\draw[white, fill=white] (-0.0625,0.095) 
	circle [radius=0.007];
	\end{tikzpicture}
	\hspace{-2mm}
}
\foreach \x in {A, ..., Z}{\expandafter\xdef\csname orcid\x\endcsname{\noexpand\href{https://orcid.org/\csname orcidauthor\x\endcsname}
			{\noexpand\orcidicon}}
}
\usepackage{authblk}

\author[1\thanks{e-mail: \tt{p.kremer@uni.lu}}]{Paul Kremer\orcidA{}}
\author[1]{Hriday Bavle\orcidB{}}
\author[1]{Jose Luis Sanchez-Lopez\orcidC{}}
\author[1,2]{Holger Voos\orcidD{}}

\affil[1]{University of Luxembourg, Interdisciplinary Center for Security Reliability and Trust (SnT), 29 Av. J.F. Kennedy, L-1855 Luxembourg, Luxembourg}
\affil[2]{University of Luxembourg, Faculty of Science, Technology and Medicine, 2 Avenue de l'Université, L-4365 Esch-sur-Alzette, Luxembourg}
\begin{document}
% ======================================================================================================
\maketitle
% ======================================================================================================
  
\begin{abstract}
Path planning is a basic capability of autonomous mobile robots. Former approaches in path planning exploit only the given geometric information from the environment without leveraging the inherent semantics within the environment. 
The recently presented \textit{S-Graphs} constructs 3D situational graphs incorporating geometric, semantic, and relational aspects between the elements to improve the overall scene understanding and the localization of the robot. 
But these works do not exploit the underlying semantic graphs for improving the path planning for mobile robots. 
To that aim, in this paper, we present \textit{S-Nav} a novel semantic-geometric path planner for mobile robots. 
It leverages \textit{S-Graphs} to enable fast and robust hierarchical high-level planning in complex indoor environments. 
The hierarchical architecture of \textit{S-Nav} adds a novel semantic search on top of a traditional geometric planner as well as precise map reconstruction from \textit{S-Graphs} to improve planning speed, robustness, and path quality. 
We demonstrate improved results of \textit{S-Nav} in a synthetic environment. 

\end{abstract}
\vspace{0.35cm}

% ======================================================================================================
\section*{License}
% ======================================================================================================
For the purpose of Open Access, the author has applied a CC-BY-4.0 public copyright license to any Author Accepted Manuscript version arising from this submission.

% % ======================================================================================================
% \section*{Acknowledgements}
% % ======================================================================================================
% This work was partially funded by the Fonds National de la Recherche of Luxembourg (FNR) under the project 17097684/RoboSAUR, by a partnership between the SnT-UL and Stugalux Construction S.A., and by the Spanish and Aragon governments (projects PID2021-127685NB-I00, TED2021-131150B-I00 and DGA FSE-T45 20R).

% ======================================================================================================
\section{Introduction}
% ======================================================================================================
Mobile robots have gained a lot of traction in recent years and have seen widespread use in different industries such as construction, mining, etc., where they are used for autonomous inspection tasks. 
To date, they are mostly teleoperated or operated semi-autonomously under the supervision of a human operator. 
Fully autonomous operation could thus significantly reduce costs, however, several technical challenges such as perception, navigation, mapping, and localization are currently detrimental to this mode of operation. 
Mobile robots should not only create meaningful maps of the environment while localizing within it but also be able to exploit these maps to perform fast and efficient planning. 

Traditionally, mobile robots build a geometric map \cite{Oleynikova2017a}, \cite{oleynikova2018sparse} of their environment using simultaneous localization and mapping techniques (SLAM) in combination with their onboard sensors \cite{Zhang2017} (e.g., LiDAR). 
Recently, we presented \textit{S-Graphs} a novel graph-based semantic SLAM that combines traditional geometric SLAM with scene graphs \cite{Bavle2022}, \cite{Bavle2022a}.
\textit{S-Graphs} extracts the topological-relational information of the environment such as wall surfaces, rooms, and doorways including the topological connections between those semantic entities enabling the robot to reason about its environment in a way humans would. 
%Though promising in terms of precise robot localization and high-level hierarchical map generation, currently \textit{S-Graphs} is validated over datasets and is not leveraged for performing more intelligent and faster path planning for mobile robots. 
\textit{S-Graphs} showed promising results in terms of precise robot localization and high-level hierarchical map generation over a variety of datasets.
However, this scene knowledge is not yet leveraged for performing more intelligent and faster path planning for mobile robots. 

To bridge this gap, we leverage the metric, semantic, and relational information in \textit{S-Graphs} for the purpose of path planning. 
We propose a novel hierarchical planner called \textit{S-Nav} which leverages the semantic layer to improve planning on the geometric layer. 
First, we perform a semantic graph search utilizing the semantic elements within the \textit{S-Graphs} to generate a sparse undirected graph of semantic elements such as rooms and doorways. 
The undirected global semantic graph is then divided into local subproblems which can be solved in parallel and pose a set of simpler problems to the underlying geometric planner. 
The main contributions of this work are:
\begin{itemize}
    \item Novel hierarchical planner called \textit{S-Nav} utilizing geometric, semantic, and relational information for faster planning. 
    \item Semantic planner for faster global plans.
    \item Semantic subproblem solver further simplifies the global plan into local subproblems for the underlying geometric planner. 
    %\item Validation of the proposed work over synthetic datasets showing improved performance in terms of time and accuracy. 
\end{itemize}

A brief summary of \textit{S-Graphs} is given in \cref{sec:s-graphs}.
\textit{S-Nav}, the novel semantic-geometric planner, is introduced in \cref{sec:s-nav}.
The main blocks are the \textit{Semantic Planner} (\cref{sec:semantic-search}), the \textit{Subproblem Solver} (\cref{sec:subproblem-solver}), and the \textit{Geometric Planner} (\cref{sec:geometric-search}).
The evaluation and results are presented in \cref{sec:results}.
This work is concluded in \cref{sec:conclusion}.

% ======================================================================================================
\section{System Overview}
\begin{figure}
    \centering
    \resizebox{0.8\linewidth}{!}{%
    \tikzset{every picture/.style={line width=0.75pt}} %set default line width to 0.75pt        

\begin{tikzpicture}[x=0.75pt,y=0.75pt,yscale=-1,xscale=1]
%uncomment if require: \path (0,262); %set diagram left start at 0, and has height of 262

%Flowchart: Process [id:dp1659276554553285] 
\draw  [fill={rgb, 255:red, 239; green, 239; blue, 239 }  ,fill opacity=1 ] (150,60) -- (590,60) -- (590,240) -- (150,240) -- cycle ;
%Flowchart: Process [id:dp4303632630512608] 
\draw  [fill={rgb, 255:red, 197; green, 215; blue, 232 }  ,fill opacity=1 ] (160,110) -- (340,110) -- (340,170) -- (160,170) -- cycle ;
%Flowchart: Process [id:dp6503587740775252] 
\draw  [fill={rgb, 255:red, 213; green, 235; blue, 192 }  ,fill opacity=1 ] (380,110) -- (580,110) -- (580,170) -- (380,170) -- cycle ;
%Flowchart: Process [id:dp08179498126760576] 
\draw  [fill={rgb, 255:red, 215; green, 244; blue, 237 }  ,fill opacity=1 ] (180,10) -- (240,10) -- (240,50) -- (180,50) -- cycle ;
%Flowchart: Process [id:dp4750613760242077] 
\draw  [fill={rgb, 255:red, 255; green, 255; blue, 255 }  ,fill opacity=1 ] (260,120) -- (330,120) -- (330,160) -- (260,160) -- cycle ;
%Flowchart: Process [id:dp5414071814084855] 
\draw  [fill={rgb, 255:red, 255; green, 255; blue, 255 }  ,fill opacity=1 ] (480,120) -- (570,120) -- (570,160) -- (480,160) -- cycle ;
%Flowchart: Process [id:dp9579397492650968] 
\draw  [fill={rgb, 255:red, 255; green, 255; blue, 255 }  ,fill opacity=1 ] (390,120) -- (460,120) -- (460,160) -- (390,160) -- cycle ;
%Flowchart: Process [id:dp08952887340218518] 
\draw  [fill={rgb, 255:red, 255; green, 255; blue, 255 }  ,fill opacity=1 ] (170,120) -- (240,120) -- (240,160) -- (170,160) -- cycle ;
%Straight Lines [id:da27772485571867955] 
\draw    (210,50) -- (210,107) ;
\draw [shift={(210,110)}, rotate = 270] [fill={rgb, 255:red, 0; green, 0; blue, 0 }  ][line width=0.08]  [draw opacity=0] (6.25,-3) -- (0,0) -- (6.25,3) -- cycle    ;
%Straight Lines [id:da6646140270845767] 
\draw    (210,80) -- (430,80) -- (430,107) ;
\draw [shift={(430,110)}, rotate = 270] [fill={rgb, 255:red, 0; green, 0; blue, 0 }  ][line width=0.08]  [draw opacity=0] (6.25,-3) -- (0,0) -- (6.25,3) -- cycle    ;
%Flowchart: Process [id:dp3495563603608397] 
\draw  [fill={rgb, 255:red, 255; green, 242; blue, 215 }  ,fill opacity=1 ] (420,190) -- (510,190) -- (510,230) -- (420,230) -- cycle ;
%Straight Lines [id:da30891638999726323] 
\draw    (417,210) -- (360,210) -- (360,140) -- (340,140) ;
\draw [shift={(420,210)}, rotate = 180] [fill={rgb, 255:red, 0; green, 0; blue, 0 }  ][line width=0.08]  [draw opacity=0] (6.25,-3) -- (0,0) -- (6.25,3) -- cycle    ;
%Straight Lines [id:da6754734584492149] 
\draw    (440,173) -- (440,190) ;
\draw [shift={(440,170)}, rotate = 90] [fill={rgb, 255:red, 0; green, 0; blue, 0 }  ][line width=0.08]  [draw opacity=0] (6.25,-3) -- (0,0) -- (6.25,3) -- cycle    ;
%Flowchart: Process [id:dp38321288954841704] 
\draw  [fill={rgb, 255:red, 255; green, 255; blue, 255 }  ,fill opacity=1 ] (620,190) -- (680,190) -- (680,230) -- (620,230) -- cycle ;
%Straight Lines [id:da6680429882871995] 
\draw    (110,140) -- (157,140) ;
\draw [shift={(160,140)}, rotate = 180] [fill={rgb, 255:red, 0; green, 0; blue, 0 }  ][line width=0.08]  [draw opacity=0] (6.25,-3) -- (0,0) -- (6.25,3) -- cycle    ;
%Straight Lines [id:da6318770282125832] 
\draw    (510,210) -- (617,210) ;
\draw [shift={(620,210)}, rotate = 180] [fill={rgb, 255:red, 0; green, 0; blue, 0 }  ][line width=0.08]  [draw opacity=0] (6.25,-3) -- (0,0) -- (6.25,3) -- cycle    ;
%Straight Lines [id:da3927283020314264] 
\draw    (650,190) -- (650,30) -- (243,30) ;
\draw [shift={(240,30)}, rotate = 360] [fill={rgb, 255:red, 0; green, 0; blue, 0 }  ][line width=0.08]  [draw opacity=0] (6.25,-3) -- (0,0) -- (6.25,3) -- cycle    ;
%Straight Lines [id:da6603216427625839] 
\draw    (490,170) -- (490,187) ;
\draw [shift={(490,190)}, rotate = 270] [fill={rgb, 255:red, 0; green, 0; blue, 0 }  ][line width=0.08]  [draw opacity=0] (6.25,-3) -- (0,0) -- (6.25,3) -- cycle    ;

% Text Node
\draw (210,29) node  [font=\footnotesize] [align=left] {\begin{minipage}[lt]{40.8pt}\setlength\topsep{0pt}
\begin{center}
S-Graphs
\end{center}

\end{minipage}};
% Text Node
\draw (295,140) node  [font=\footnotesize] [align=left] {\begin{minipage}[lt]{44.2pt}\setlength\topsep{0pt}
\begin{center}
Semantic\\Planner
\end{center}

\end{minipage}};
% Text Node
\draw (525,140) node  [font=\footnotesize] [align=left] {\begin{minipage}[lt]{61.2pt}\setlength\topsep{0pt}
\begin{center}
Sampling-base\\ Planner
\end{center}

\end{minipage}};
% Text Node
\draw (425,140) node  [font=\footnotesize] [align=left] {\begin{minipage}[lt]{40.8pt}\setlength\topsep{0pt}
\begin{center}
Global\\Map
\end{center}

\end{minipage}};
% Text Node
\draw (530,100) node  [font=\scriptsize] [align=left] {\begin{minipage}[lt]{68pt}\setlength\topsep{0pt}
\begin{flushright}
Geometric Planner
\end{flushright}

\end{minipage}};
% Text Node
\draw (205,140.54) node  [font=\footnotesize] [align=left] {\begin{minipage}[lt]{47.6pt}\setlength\topsep{0pt}
\begin{center}
Graph\\Structure
\end{center}

\end{minipage}};
% Text Node
\draw (295,100) node  [font=\scriptsize] [align=left] {\begin{minipage}[lt]{61.2pt}\setlength\topsep{0pt}
\begin{flushright}
Semantic Planner
\end{flushright}

\end{minipage}};
% Text Node
\draw (400,71.02) node  [font=\scriptsize] [align=left] {\begin{minipage}[lt]{40.8pt}\setlength\topsep{0pt}
\begin{flushright}
Graph Data
\end{flushright}

\end{minipage}};
% Text Node
\draw (125,133) node  [font=\footnotesize] [align=left] {\begin{minipage}[lt]{34pt}\setlength\topsep{0pt}
\begin{center}
Query
\end{center}

\end{minipage}};
% Text Node
\draw (570,250) node  [font=\footnotesize] [align=left] {\begin{minipage}[lt]{27.2pt}\setlength\topsep{0pt}
\begin{flushright}
S-Nav
\end{flushright}

\end{minipage}};
% Text Node
\draw (465,210) node  [font=\footnotesize] [align=left] {\begin{minipage}[lt]{47.6pt}\setlength\topsep{0pt}
\begin{center}
Subproblem\\Solver
\end{center}

\end{minipage}};
% Text Node
\draw (650,210) node  [font=\footnotesize] [align=left] {\begin{minipage}[lt]{28.9pt}\setlength\topsep{0pt}
\begin{center}
Robot
\end{center}

\end{minipage}};
% Text Node
\draw (550,220) node  [font=\scriptsize] [align=left] {\begin{minipage}[lt]{54.4pt}\setlength\topsep{0pt}
\begin{center}
Global Path
\end{center}

\end{minipage}};
% Text Node
\draw (355,220) node  [font=\scriptsize] [align=left] {\begin{minipage}[lt]{88.4pt}\setlength\topsep{0pt}
Semantic-geometric Path
\end{minipage}};
% Text Node
\draw (300,20) node  [font=\scriptsize] [align=left] {\begin{minipage}[lt]{54.4pt}\setlength\topsep{0pt}
\begin{center}
Sensor Data
\end{center}

\end{minipage}};
% Text Node
\draw (400,180) node  [font=\scriptsize] [align=left] {\begin{minipage}[lt]{40.8pt}\setlength\topsep{0pt}
Subproblem
\end{minipage}};
% Text Node
\draw (540,180) node  [font=\scriptsize] [align=left] {\begin{minipage}[lt]{54.4pt}\setlength\topsep{0pt}
Subpath
\end{minipage}};

\end{tikzpicture}%
    }
    \caption{Overview of \textit{S-Nav}. A \textit{query} (e.g., 'go from current position to the kitchen') is first handled by the \textit{Semantic Planner} which provides an initial high-level semantic-geometric path based on the graph structure obtained from \textit{S-Graphs}. Next, in the \textit{Subproblem Solver}, the semantic-geometric path is subdivided into smaller, easier problems that are individually solved by the \textit{Geometric Planner}. The individual paths between the subproblems are then reassembled into the final high-level path that is passed to the robot.}
    \label{fig:system-overview}
\end{figure}
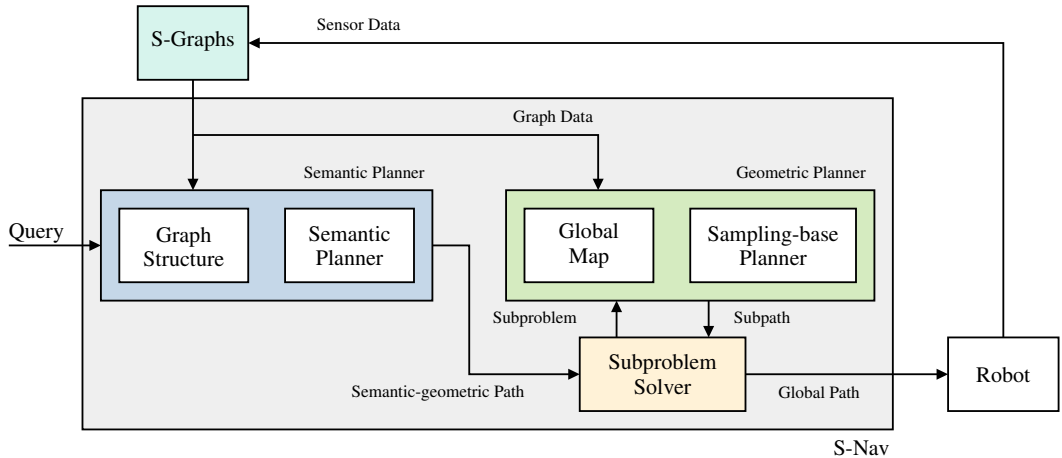

% ======================================================================================================
The complete system architecture is shown in \cref{fig:system-overview}.
\textit{S-Nav} builds on top of \textit{S-Graphs} and utilizes it as its main data source.
\textit{S-Nav} itself is composed of the \textit{Semantic Planner}, the \textit{Subproblem Solver}, and the \textit{Geometric Planner}.
A path query is first handled first by the \textit{Semantic Planner} whose output serves as a rough initial guess that cascades into the \textit{Geometric Planner} via the \textit{Subproblem Solver}.

% ======================================================================================================
\subsection{Situational Graphs (S-Graphs)}
% ======================================================================================================
\label{sec:s-graphs}
\textit{S-Graphs} is an optimizable graph structure built using online measurements such as LiDAR data or markers \cite{Bavle2022}, \cite{Bavle2022a}, \cite{Shaheer2023a}.

The graph structure consists of five layers that are summarized as:

\textbf{Keyframes Layer}: Composed by the robot's pose ${}^ Mx_{R_i} \in SE(3)$ constrained by the the robot's odometry measurements.

\textbf{Walls Layer}: Each room is composed of four planes extracted from onboard sensor measurements. 
They are constrained using pose-plane constraints.

\textbf{Room Layer}: A room is formed by its four planes constrained by a cost function consisting of the room center ${}^M\mathbf{p}_i \in \mathbb{R}^2$ and $w_i$ the distance between the opposite planar pairs.

\textbf{Floor Layer}: A floor is a collection of rooms optimized analogously to rooms by extracting the largest distance between the opposite planar pairs.

\textbf{Doorway Layer}: A doorway marks the physical, traversable connection between two rooms defined by a center point ${}^M\mathbf{d}_i \in \mathbb{R}^2$ and a width $r_i$, and is constrained by the physical distance between the two rooms it connects.

\textit{S-Graphs} serves as the main source of information for \textit{S-Nav}.
Therefore, this work makes extensive use of the room and doorway layers. 
The presented architecture can, however, easily be expanded to include multiple floors and other semantic entities (e.g., objects).

% ======================================================================================================
\section{S-Nav}
% ======================================================================================================
\label{sec:s-nav}
\textit{S-Nav} is our novel hierarchical semantic-geometric planning solution that combines \textit{S-Graphs} with an informed geometric planner.
Our solution provides the following benefits over traditional, purely geometric planners:
\begin{itemize}
    \item The geometric search can greatly profit from a rough initial guess provided by the semantic layer by constraining the areas the planner can visit and by providing subgoals toward the final goal.
    \item A query in natural form, e.g., 'go from here to the kitchen', can easily be mapped to a semantic-geometric problem.
    \item Handling forbidden areas such as closed doors or rooms that should not be traversed is trivial on the semantic layer, whereas it would require map changes on the geometric layer. 
    Similarly, if a doorway is detected as untraversable, replanning is very fast as the doorways node can easily be disconnected from the graph.
\end{itemize}

\begin{figure}
    \centering
    \resizebox{0.75\linewidth}{!}{%
    \input{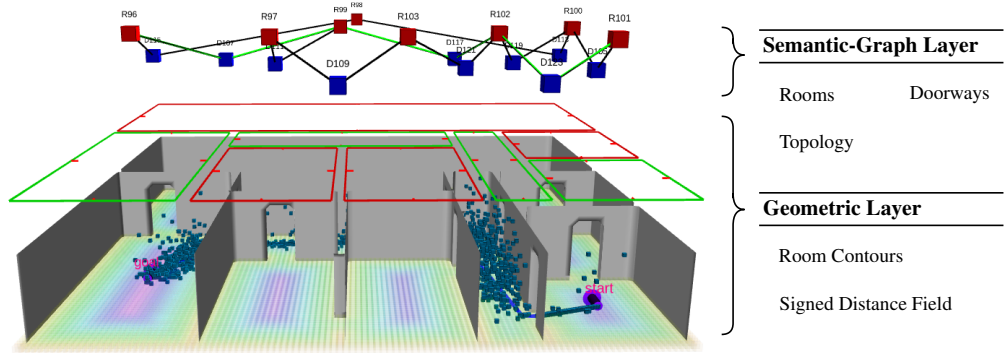}%
    }
    \caption{The semantic-graph layer is an undirected topological graph of rooms and doorways extracted from features segmented by \textit{S-Graphs}. The results of this first layer cascade into the geometric layer which is formed by OMPL and Voxblox. The contour layer reduces the valid state space of the geometric planner to do informed decisions on where to sample. The semantic path is highlighted in green together with the room contours that have to be traversed along the path.}
    \label{fig:snav-layers}
\end{figure}

The structure of \textit{S-Nav} is depicted in \cref{fig:system-overview} whilst the different layers are visualized in \cref{fig:snav-layers}.
Its main parts are formed by the \textit{Semantic Planner} that cascades into the \textit{Geometric Planner} via the \textit{Subproblem Solver} (SPS).
The final path is then passed to the robot, potentially via additional layers such as local planning, trajectory generation, and motion control.

% ======================================================================================================
\subsection{Semantic Search}
\label{sec:semantic-search}
% ======================================================================================================

The scene graph structure of \textit{S-Graphs} encodes a high-level representation of the environment the robot is operating in.
Herein, this scene graph is converted into an undirected graph connecting the semantic elements of the scene.
The connections (edges) between the elements have an associated cost, i.e., for the room-to-doorway connections, a cost
\begin{align}
    c_{dr} = \lVert {}^M\mathbf{p}_i - {}^M\mathbf{d}_i \rVert^2 + p_{d}
\end{align}
is assigned, consisting of the distance between the center point of the room and the associated doorway, plus a fixed penalty for doorway crossing.
The fixed penalty $p_{d}$ can be used to prefer a slightly longer path with fewer (potentially closed) doorways.

Generally, the graph is sparse, featuring only a small number of nodes and edges.
As such, a shortest-path search using, e.g., A* is virtually free compared to a full search on the geometric layer. 

For a given \textit{query} '$\mathbf{p}_s \rightarrow \mathbf{p}_g$' (read: from $\mathbf{p}_s$ to $\mathbf{p}_g$), the semantic planner provides a \textit{solution} of type
\begin{align}
\begin{split}
   \mathbf{p}_s \rightarrow \mathbf{d}_k \rightarrow \mathbf{d}_{k+1} \rightarrow \cdots \rightarrow \mathbf{d}_{k+n} \rightarrow \mathbf{p}_g \\
   \textbf{via:}\hspace{0.1cm} F_R = R_s \cup R_{i} \cup \dots  \cup  R_{i+n} \cup R_g,\hspace{0.2cm} F_R \subseteq S \subset\mathbb{R}^3
   \label{eq:semantic-search}
\end{split}
\end{align}
where $\mathbf{p}_s, \mathbf{p}_g$, $R_s, R_g$ are the start and goal positions and rooms, $\mathbf{d}_{k}$ the doorway center points to traverse along the route, $R_{(.)}$ are the rooms, i.e., the free space the robot has to pass through to reach its destination, $S$ is the state space limited by the bounding box of the map, and $F_R$ is the reduced free space obtained from the semantic planner that is passed to the geometric planner.
By restricting the geometric planner to $F_R$, its sampler can be much smarter about the placement of the samples.

% ======================================================================================================
\subsection{Semantic Search with Subproblems}
\label{sec:subproblem-solver}
% ======================================================================================================
The global problem in \cref{eq:semantic-search} can further be simplified into a set of local subproblems that can be solved in parallel and, individually, pose a simpler problem to the geometric planner:
\begin{equation}
\begin{alignedat}{3}
   &1: \mathbf{p}_s &&\rightarrow \mathbf{d}_{k}, &&\phantom{0}\textbf{via: } R_{s\phantom{+0}} \\
   &2: \mathbf{d}_{k} &&\rightarrow \mathbf{d}_{k+1}, &&\phantom{0}\textbf{via: } R_{i+1} \\
   &3: \mathbf{d}_{k+1} &&\rightarrow \mathbf{d}_{k+2}, &&\phantom{0}\textbf{via: } R_{i+2} \\
   &  &&\cdots &&\\
   &n: \mathbf{d}_{k+n} &&\rightarrow \mathbf{p}_g, &&\phantom{0}\textbf{via: } R_{g\phantom{+0}}
\end{alignedat}
\end{equation}

Therefore, akin to informed geometric planners (e.g., informed RRT* \cite{Gammell2014}), herein the semantic planner adds an additional layer of information that the geometric planner can profit from to find a solution faster.
The subproblems are solved by the \textit{Subproblem Solver} in conjunction with the geometric planner. 
The resulting individual path segments are joined to a final, global path.
If the resulting path requires updating, e.g., due to a blocked path, the \textit{Subproblem Solver} can efficiently reevaluate the changed or newly created subproblems.

% ======================================================================================================
\subsection{Global Map}
\label{sec:global-map}
% ======================================================================================================
Instead of relying on raw sensor readings, \textit{S-Nav} features a global map reconstruction module that builds an accurate global map from \textit{S-Graphs} data, which in turn is generated on the fly by the robot, resp. provided to the robot if the environment is already fully mapped.
The global map is kept relatively simple, i.e., not featuring obstacles as this problem is more effectively handled by the reactive planner on a local map.

\textit{S-Graphs} provides the planes associated with each room, including the doorways that mark the connection between two rooms.
In the first step, for each room, the vertical planes (walls) are converted to a closed 2D contour which encompasses the free space within a room.
Room contours serve two purposes: 
\textbf{First}, to restrict the geometric planner's sampler to sample only in areas that effectively contribute to the final path.
\textbf{Second}, to build an optimistic, clutter-free (yet accurate), signed distance field representation of the physical environment that forms the basis for the geometric planner.

Doorways of a certain width and located at a certain point are added between the two closest walls of the two associated rooms.
Just like contours, they are also part of the signed distance field generation process.

% ======================================================================================================
\subsection{Geometric Search}
\label{sec:geometric-search}
% ======================================================================================================

\begin{figure*}[t]
    \centering
    \resizebox{1.0\linewidth}{!}{%
    \input{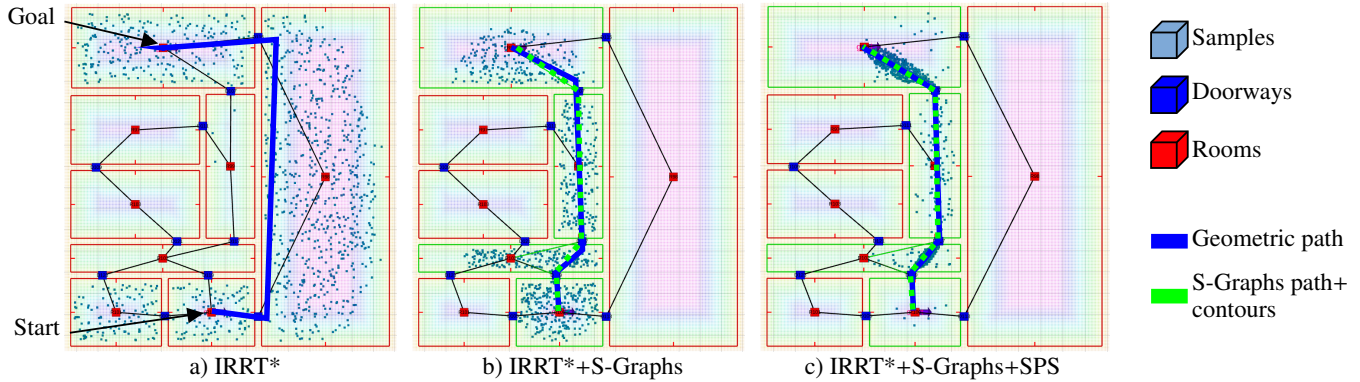}%
    }
    \caption{a) The naïve approach using IRRT* creates an excessive amount of samples and yet ends up with a suboptimal path. b) Restricting IRRT* to sample within the rooms that are part of the optimal path greatly enhances the solution with fewer samples. c) Restricting IRRT* to sample within the rooms parts of the solution and decomposing the overall problem in subproblems yields the best result (least amount of samples) as it effectively exploits the rapid convergence of IRRT* for the resulting simpler problems.}
    \label{fig:snav-planning}
\end{figure*}

The geometric search within \textit{S-Nav} features state-of-the-art geometric planners provided by the \textit{Open Motion Planning Library} (OMPL).
Within \textit{S-Nav} we use sampling-based planners (e.g., PRM, RRT, IRRT*), that create random (sometimes with a heuristic) samples within the valid bounds of the state space.
A priori, for any given problem, the whole global map has to be considered.
Therefore, a large number of samples is required to find the optimal path.
Constraining the sampler to sample within the rooms that have to be visited along the semantic path, greatly enhances the convergence rate of the planner as no samples are wastefully created in areas that are of no interest.
Furthermore, using the SPS that decomposes the global problem into a set of local problems effectively exploits the rapid convergence of certain planning algorithms (e.g., IRRT*) for the resulting simpler type of problem.
The problem is illustrated in \cref{fig:snav-planning}.

% ======================================================================================================
\section{Evaluation}
% ======================================================================================================
\label{sec:results}

% ======================================================================================================
\subsection{Methodology}
% ======================================================================================================
A synthetic map ($\SI{17}{\meter}\cross\SI{15}{\meter}$) with $8$ rooms and $10$ doorways was created and passed to the recently presented \textit{iS-Graphs} \cite{Shaheer2023a}, an \textit{S-Graphs} extension that supports architectural (BIM) data.
Within this environment, the three cases (IRRT*, IRRT*+S-Graphs, and IRRT*+S-Graphs+SPS) shown in \cref{fig:snav-planning} were benchmarked by performing $1000$ queries for each.
As OMPL termination criteria, a timeout of \SI{0.1}{\second} was specified.
The timeout is equally divided over all subproblems for the test series involving the SPS.
Recorded were the number of samples created within the allocated time as well as the final path length.
The measurements were performed on a workstation equipped with an Intel Core i9-11950H.

% ======================================================================================================
\subsection{Results and Discussion}
\begin{figure}
    \centering
    \includegraphics[width=0.4\linewidth]{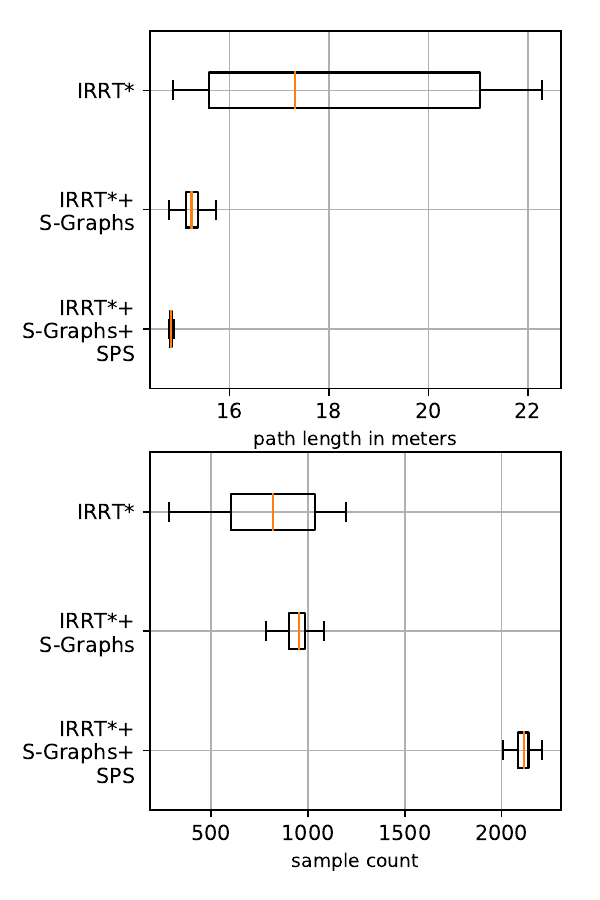}
    \caption{Results: The semantic planner greatly improves the consistency of the results by reliably finding the shorter path compared to plain IRRT*. The solution involving IRRT*, \textit{S-Graphs}, and the SPS was able to create significantly more samples in the allocated time.}
    \label{fig:results}
\end{figure}

% ======================================================================================================
The results are given in \cref{fig:results}.
It is clear that IRRT* alone delivers the most inconsistent results with the widest spread.
On average, it also had the least number of samples generated.
Restricting the sampled regions with the \textit{S-Graphs} knowledge, significantly improved the consistency of the results.
Further, using \textit{S-Graphs} and the SPS in combination with IRRT* yielded consistently the shortest path, and was also able to generate significantly more samples.
The higher number of samples is caused by the comparatively cheaper state and motion validity checks based on the contours rather than the signed distance field alone.

% ======================================================================================================
\section{Conclusion} 
% ======================================================================================================
\label{sec:conclusion}
Leveraging the geometric-semantic knowledge contained in \textit{S-Graphs} for planning can greatly enhance the performance of the underlying geometric planner.
Herein, we presented \textit{S-Nav}, a novel semantic-geometric planner that features a hierarchical planner architecture that showed to significantly improve planning speed, resp., the consistency of the generated paths within a given timeframe.
Furthermore, we showed that decomposing the global problem into a set of local problems can be used to effectively leverage the rapid convergence of (informed) sampling-based planners.

\bibliographystyle{plainnat}
\bibliography{mybib}
\end{document}